# Potential of Combined Learning Strategies to Enhance Energy Efficiency of Spiking Neuromorphic Systems

Shiri Sichani, Ali
University of Missouri-Columbia
Columbia, Missouri, USA

Sai Kankatala
University of Missouri-Columbia
Columbia, Missouri, USA

## ABSTRACT

Ensuring energy-efficient design in neuromorphic computing systems necessitates a tailored architecture combined with algorithmic approaches. This manuscript focuses on enhancing brain-inspired perceptual computing machines through a novel combined learning approach for Convolutional Spiking Neural Networks (CSNNs). CSNNs present a promising alternative to traditional power-intensive and complex machine learning methods like backpropagation, offering energy-efficient spiking neuron processing inspired by the human brain. The proposed combined learning method integrates Pair-based Spike Timing-Dependent Plasticity (PSTDP) and power-law-dependent Spike-timing-dependent plasticity(STDP) to adjust synaptic efficacies, enabling the utilization of stochastic elements like memristive devices to enhance energy efficiency and improve perceptual computing accuracy. By reducing learning parameters while maintaining accuracy, these systems consume less energy and have reduced area overhead, making them more suitable for hardware implementation. The research delves into neuromorphic design architectures, focusing on CSNNs to provide a general framework for energy-efficient computing hardware. Various CSNN architectures are evaluated to assess how less trainable parameters can maintain acceptable accuracy in perceptual computing systems, positioning them as viable candidates for neuromorphic architecture. Comparisons with previous work validate the achievements and methodology of the proposed architecture.

## CCS CONCEPTS

- **Hardware** → VLSI design; Emerging devices;
- **Computer systems organization** → Neural networks;
- **Computing methodologies** → Learning linear models.

## 1 INTRODUCTION

Spiking Neural Networks (SNNs) are at the forefront, replicating brain signals through discrete "spikes" to capture the temporal dynamics of information processing [1]. The availability of energy-efficient chip sets for neuromorphic computing remains limited. The cognitive and perceptual computing challenges posed by von Neumann architecture have prompted the exploration of novel computing paradigms. Traditional processors struggle with speed limitations and data transfer inefficiencies, driving the need for more efficient processing methods. Recent advancements in computing, fueled by the demands of big data analysis, have led to multidisciplinary research in intelligent systems for cloud and mobile computing applications [2]. Heterogeneous computing architectures enable the computing system to cope with big data processing [3]. Brain-inspired machine learning methods have paved the way for creating novel neuromorphic systems. Within the last two decades, various large projects have been proposed to develop neuromorphic processing systems such as FACETS [4], SpiNNaker [5], TrueNorth[6], Neurogrid[7], and Loihi Intel neuromorphic chip [8].

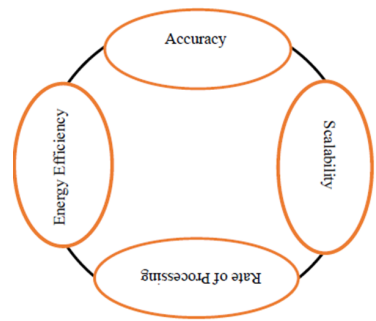

**Figure 1: The significant factors in the design of architecture**

However, the architectures mentioned earlier have not been entirely successful in addressing the bottleneck of von Neuman architecture, as they relied on multicore processing. In the era of the Internet of Things (IoT), neuromorphic event-driven systems offer an efficient solution for lightweight processing. The improvement in computing systems requires compromising between computational hardware-oriented factors demonstrated in Figure 1. Neuromorphic computing is becoming a more realistic path in the search for more effective computational models. These systems provide significant gains in processing power and energy efficiency over conventional architectures by emulating the neural architecture and functions of the human brain. Spiking Neural Networks (SNNs), which use discrete "spikes" to process information dynamically and capture the temporal nuances of neural activity, are at the center of this endeavor. Even with significant advancements in the field, SNNs' practical application potential is still mostly unrealized because of the difficulties in training and integrating them with current hardware.

The significant trade-off among computing circuit factors demonstrated in Figure 1 hinders designers from delivering energy-efficient computing hardware based on conventional computer architectures, known as von Neumann architecture, this challenge is crucial particularly for big data processing. Alternative computing circuit design approaches are related to changes in architectural structure and a paradigm shift in computing methodologies, such as neuro-inspired computing, which can provide ultra-energy-efficient computing circuits and systems, particularly when implemented using memristive devices. Brain-inspired computing circuits are developed based on two platforms. 1- conventional NNs and 2-SNNs. [9]. The conventional NNs process the data using heavy computational backpropagation for synaptic weight updates. Applying traditional NNs methods for computing circuits can reduce the energy consumption of the circuits, but they still consume significant energy[10]. The design of computing circuits based on spiking neural networks, specifically under a spiking time-dependent plasticity (STDP) regime for synaptic updates, provides the opportunity to emulate brain behavior for conceptual processing and ultimately dramatically reduce the energy consumption of computing circuits. Due to the wide range and complexity of the neuron models, brain-inspired computing at the circuit level does not necessarily provide the required energy efficiency. Selecting an appropriate neuron model and synaptic update laws that align with the flexibility of existing technology in circuits and devices is crucial for designing efficient brain-inspired computing. A successful methodology for developing neuro-inspired hardware should focus on achieving a balanced trade-off. This involves harmonizing the complexity of the overall architecture with the simplicity of its components. By doing so, it ensures that the system is both efficient and scalable while maintaining ease of manufacturing and cost-effectiveness. Recent developments in memristive devices for neuromorphic computing have steered the attention of researchers to the development of computing circuits and systems based on emerging devices that often demonstrate memristive phenomena in their characteristics. The intricate learning behavior of memristive devices, which do not adhere strictly to established synaptic update patterns in neuroscience, sparked the thinking of employing a combined learning strategy. This strategy aims to determine its effectiveness in creating energy-efficient Convolutional Spiking Neural Networks (CSNNs) that boast fewer trainable parameters relative to other CSNN variants. This research utilizes the Leaky Integrate-and-Fire (LIF) neuron model to emulate neuron behavior by considering an additional coefficient in the first differential equation of the neuron. The LIF model emerges as a straightforward and extensively validated representation of primate visual cortex neurons. Stimulation of the LIF model can be achieved using Poisson spike train pulses, which are considered the most realistic pulse regime by neuroscientists for replicating the brain's stochastic pulse generation [11]. This paper deploys combined learning for some architectures of CSNNs that merge the hierarchical, feature-detection capabilities of traditional Convolutional Neural Networks (CNNs) with the dynamic, event-driven nature of SNNs. The proposed CSNN architecture leverages a unique combined learning approach that integrates both Pair-based STDP and power-law-dependent STDP to enhance synaptic efficacies performance. The successful integration of CSNN in achieving energy-efficient architecture not only paves the way for alogrithmic architectural in hardware implementation of stochastic elements like memristive devices but also enhances the system's overall energy efficiency and computational accuracy. By addressing the challenges of synaptic efficiency and system integration, our work sets the stage for the next generation of neuromorphic computing systems.

## 1.1 Correlation between Energy Efficiency and Number of Synaptic Connections

In the design of neuromorphic chips based on different structures [6][12] [8], synaptic connections account for the most significant portion of circuit activity during training. This is because the synaptic connections must be updated in response to incoming events. From a digital design perspective, the dynamic power consumption of circuits within the synaptic connections is substantial, even though alternative technologies based on emerging non-volatile memories can reduce energy consumption. However, even these alternative technologies consume most significant part of energy consumption when exploited as synaptic connections[13][14]. The high power consumption of synaptic connections during training arises from the frequent updates required to adjust the connection strengths based on the input data and learning algorithm. Each update necessitates reading the current synaptic weight, modifying it according to the learning rule, and writing the updated weight back to memory. These read-modify-write operations can be energy-intensive, particularly when performed in parallel across numerous synaptic connections [15]. While emerging non-volatile memory technologies, such as resistive RAM (RRAM) or phase-change memory (PCM), offer lower static power consumption compared to traditional SRAM or DRAM, their dynamic power consumption during write operations can still be significant [16][17]. Additionally, the endurance and reliability of these emerging memories may be a concern when subjected to the high number of write cycles required for training large neural networks [18].To mitigate the energy consumption of synaptic connections, various techniques have been explored, including:

1-Employing analog or mixed-signal circuits that can directly perform synaptic weight updates with lower energy consumption compared to digital counterparts [19]. 2-Implementing sparse weight updates, where only the synaptic connections with significant changes are updated, reducing the overall number of write operations [20]. 3-Exploring novel memory technologies with improved energy efficiency, endurance, and density for synaptic weight storage and updates [21][22].

This research recommend combined learning to reduce the number of synaptic connections while it keeps the accuracy of the computing. The combination of combined learning with our proposed method provide the chance to hugely enhance the energy efficiency of the computing system. Ultimately, minimizing the energy consumption of synaptic connections is crucial for enabling efficient and scalable neuromorphic computing systems capable of handling large-scale neural network training and inference tasks [23].

## 2 CONVOLUTIONAL SPIKING NEURAL NETWORKS

CSNNs represent an evolution of traditional CNNs in the realm of spiking neural networks, merging the power of both convolutional architectures and the temporal processing capabilities inherent in spiking neurons. Figure 2 demonstrates a CSNN which contains four convolutional layers and one max pooling layer. These typical networks extend the functionalities of CNNs by introducing temporal aspects, allowing for the encoding of information not just in the strength of connections but also in the timing of spikes.

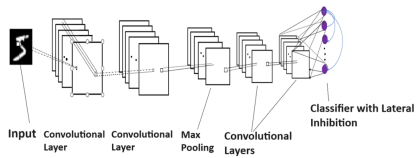

**Figure 2: General Structure of CSNN**

### 2.1 Significance and Applications of CSNNs

The convolutional structure, a hallmark of CNNs, remains a foundational aspect of CSNNs. It involves hierarchical layers that learn hierarchical representations from input data, enabling effective feature extraction and pattern recognition. However, in CSNNs, each neuron communicates through spikes, transforming the network into an event-based system, enabling precise timing information to be embedded in the information flow[24][25]. CSNNs hold significant promise for handling time-sensitive data, such as in real-time event recognition, sensory processing, and applications involving spatiotemporal patterns. Their ability to process information with temporal precision, coupled with the capability to perform convolutional operations, makes them appealing for tasks requiring both spatial and temporal understanding. As the field progresses, CSNNs stand as a bridge between the structured architecture of CNNs and the biological realism of spiking neural networks, offering a potential path toward more efficient, biologically plausible, and temporally precise artificial intelligence systems[26] Research in CSNN architectures has seen significant developments, focusing on optimizing network structures, learning algorithms, and spike-based computations. An algorithm is introduced to enable backpropagation for training CSNNs, addressing the challenges of gradient descent in spiking networks[27]. Furthermore, in TrueNorth, a hardware platform leveraging CSNNs for efficient neuromorphic computing, showcasing their potential in real-time, low-power applications[28]. The applications of CSNNs span various domains, including computer vision, robotics, and neuromorphic computing. In computer vision, CSNNs have demonstrated superior performance in object recognition tasks, utilizing their ability to process temporal information and handle spatiotemporal patterns more effectively than traditional CNNs. Neuromorphic computing applications, such as event-based sensors and brain-machine interfaces, benefit from CSNNs' event-driven processing and energy efficiency. Due to provided reasons combined learning is tested for CSNNS.

### 2.2 Integration of Hardware Design with CSNN

The concept CSNNs can be seamlessly integrated with VLSI hardware for neuromorphic computing and memristive crossbars. By leveraging the unique properties of emerging memristive devices, CSNNs can be efficiently implemented in hardware, unlocking their full potential for energy-efficient and biologically plausible computing.

Memristive crossbar arrays, composed of memristive devices arranged in a crossbar structure, offer an ideal platform for realizing the spiking behavior and weight updates required by CSNNs. These crossbars can implement the convolutional kernels and perform the necessary computations using the inherent properties of memristors, which can mimic the behavior of biological synapses and neurons [6]. In a VLSI implementation of CSNNs using memristive crossbars, the convolutional layers can be mapped onto these crossbar arrays, where the memristors act as programmable resistive elements representing the synaptic weights. The input spikes propagate through the crossbar, with the memristors modulating the signals based on their programmed resistances, effectively performing the convolution operations. The spiking nature of CSNNs can be realized by integrating the crossbar arrays with spiking neuron circuits, which can generate and transmit spikes based on the integrated input currents from the crossbars. These spiking neuron circuits can be designed using analog or mixed-signal VLSI circuits, mimicking the behavior of biological neurons and enabling the spatiotemporal processing capabilities of CSNNs[29]. Furthermore, the learning mechanisms of CSNNs, such as Spike-Timing-Dependent Plasticity (STDP), can be implemented by modulating the resistances of the memristors in the crossbar arrays based on the relative timing of pre-synaptic and post-synaptic spikes. This allows the CSNN to adapt its weights and learn from the input data in a biologically plausible manner.By combining the computational power of CSNNs with the energy-efficient and highly parallel nature of memristive crossbars, VLSI neuromorphic hardware can facilitate the realization of efficient and scalable spiking neural networks for various applications, such as real-time event recognition, sensory processing, and spatio-temporal pattern analysis [30].

### 2.3 Neuron Dynamics Model

Conventional neural networks exploit activation functions to induce nonlinearity effects in neural networks and prepare them for classification and pattern recognition. The neurons' dynamic computational model in SNNs provides much flexibility to learn and unlearn the input features[31][32][33]. The neuron's LIF model is compatible with analog-VLSI implementation to deploy analog systems for self-learning[34][35]. The LIF neuron can be realized with different dynamical parameters and provide various STDP learning to synapses[36][37]. And physical architecture has been demonstrated in Figure 3. All layers of the proposed model are populated with LIF neurons. The proposed architecture employs a computational dynamics model with a different adaptive coefficient, as shown in Equation 1:

$$\tau_{\text{mem}} \frac{dV_{\text{mem}}}{dt} = -\mu V_{\text{mem}} + \sum_j W_{j,i} \delta_j (t - t_i) \quad (1)$$

where Vmem is the neuron's membrane potential, and μ is the adaptive coefficient that gives competitive learning to the neurons. The Wj,i is the synaptic weight between the jth pre-synaptic neuron and the ith post-synaptic neuron.

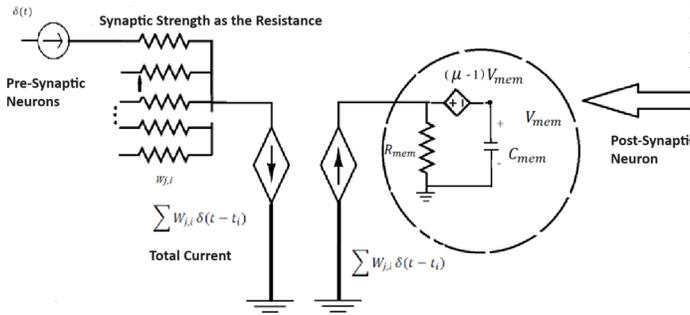

**Figure 3: Pre-synaptic neuron functions**

## 3 UNSUPERVISED COMBINED STDP LEARNING IN CONVOLUTIONAL LAYERS

The synaptic strength performs as a gain between pre and post-neurons. The STDP is a biological-inspired method for learning in SNN [32][33][38]. During network learning, the spike flows move from the input layer to the successive layers. Each post-neuron integrates the spike flows in the form of membrane voltage through synaptic connections. If the membrane voltage exceeds a certain threshold, then the post-neuron fires and phenomenological superimposition of pre and post-synaptic spikes update synaptic efficacies. Potentiation and depression of the synaptic efficacies occur using combined learning. The combination of Pair-based STDP (PSTDP) and power-law-dependent STDP boosts the synapse stochasticity and enhances perceptual computing accuracy. The biological significance of PSTDP has been proved through computational models and various circuit implementations[39]. The synaptic strength performs as a gain between pre and post-neurons. The synaptic update in pair-based STDP is given by Equation ??.

$$\Delta w_{\text{PSTDP}} = \begin{cases} \Delta W^+ = \mu^+ e^{-\left(\frac{t_{\text{post}} - t_{\text{pre}}}{\tau^+}\right)} & \text{if } t_{\text{post}} > t_{\text{pre}} \\ \Delta W^- = -\mu^- e^{+\left(\frac{t_{\text{post}} - t_{\text{pre}}}{\tau^-}\right)} & \text{if } t_{\text{post}} < t_{\text{pre}} \end{cases} \quad (2)$$

where $t_{\text{post}}$ and $t_{\text{pre}}$ are instants of pre and post-synaptic spikes, respectively. $\tau^+$ and $\tau^-$ are the time constants of depression and potentiation, respectively, while $\mu^+$ and $\mu^-$ represent the learning and unlearning rates. Finally, the depression and potentiation of synaptic strength using PSTDP are governed by the timing difference of pre and post-synaptic spikes and their ordering. The combination of the PSTDP and the power-law weight-dependent model adds up multiplicativity and additivity properties. And the Equation 3 explains about the power-law weight-dependent model[40][41]:

$$\Delta w = \lambda \times \left[ e^{\left(\frac{t_{\text{post}} - t_{\text{pre}}}{\tau}\right)} - \theta_{\text{offset}} \right] [W_{\text{max}} - w]^\mu \quad (3)$$

where $\lambda$ is the learning rate, $t_{\text{post}}$ and $t_{\text{pre}}$ are, respectively, instants of pre and post-synaptic spikes. $\tau$ is the time constant of the learning, $W_{\text{max}}$ is the maximum weight of the learning, and $\mu$ is the parameter to control the weight dependency of the learning. The $\theta_{\text{offset}}$ is a threshold level to determine the depression and potentiation of the synaptic weight. In combined learning, the learning window is the linear combination of different learning rules.

$$\Delta w = \Delta w_{\text{PSTDP}} + \Delta w_{\text{Power-law}} \quad (4)$$

Here, the combined learning as the simple paradigm demonstrates stochastic synaptic devices' significance in neuromorphic systems performance. The synaptic devices can be implemented using CMOS, BiCMOS, or memristive systems.

### 3.1 Data Preprocessing

The efficiency and accuracy of the proposed combined learning method were assessed using standard datasets, including the MNIST-Digit dataset for training and testing on 60,000 and 10,000 samples, respectively, and the Caltech dataset, containing images of faces and motorbikes. Preprocessing of the Caltech dataset involved converting the images to grayscale and applying a Laplacian of Gaussian (LoG) filter to enhance pixel contrast and edge detection, followed by resizing to 40x40 pixels. By evaluating the combined learning method on both datasets, researchers aimed to comprehensively analyze its efficiency and robustness across varying complexities of data, providing valuable insights into its effectiveness for real-world applications.

### 3.2 Training of CSNN

The convolutional window collects the spikes from the preceding layer. A larger convolutional window often collects more general features and, regarding the network topology, potentially enhances the recognition accuracy. Figure 4 demonstrates the flow of spikes from the convolutional kernel the Data classification functionality in spiking deep convolutional neural networks (DCNN) versus hyperparameters cannot be formulated explicitly, although it is tractable through simulation. The pooling layer is populated with leaky-integrate and fire-neurons, and each neuron in the neuronal channel operates a max-pooling operation over a 2×2 kernel. The pooling window sweeps the preceding convolutional layers with a stride of 2. This operation shrinks the size of thesuccessive layers. Due to control data redundancy, the max operation exploits delay units, counts the neurons' spikes inside the pooling window in a tiny time interval [10ms-20ms], and lets the pooling neuron propagate the spikes with a maximum frequency that it receives. Max-pooling operation regarding spike counts increases data efficiency and property of translation invariance. The learning of the final layer exploits the supervised STDP learning. The input patterns are preprocessed using Laplacian of Gaussian (LoG) prior to Poisson spike encoding. Figure 5 demonstrates the training method for the proposed CSNN.

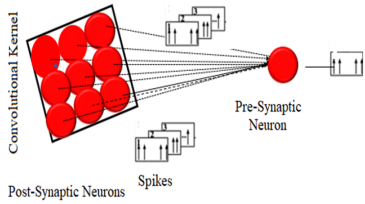

**Figure 4: Illustration of convolutional Kernel**

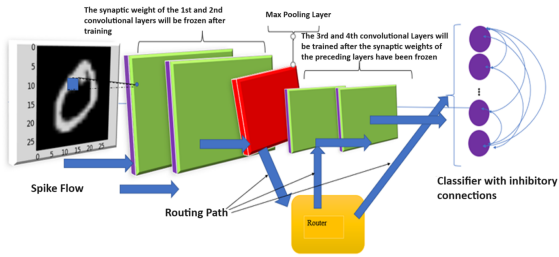

**Figure 5: Illustration of training method in proposed combined learning over SCNN. Two successive convolutional layers have been trained concurrently.**

### 3.3 Test of the Combined Learning with MNIST Dataset

In our study, we employed the Poisson spike regime to evaluate the effectiveness of combined learning in Spiking Convolutional Neural Networks (SCNNs). We investigated various SCNN architectures, including configurations denoted as 1C-1S-FC, 2C-1S-FC, and 2C-1S-2C-FC, where 'C', 'S', and 'FC' represent convolutional layers, spatial pooling layers, and fully connected layers, respectively. To gauge the performance of combined learning within SCNNs and compare it with recent research[9][33][42].

We utilized the MNIST digits dataset, which underwent preprocessing using a Laplacian of Gaussian (LoG) filter before being input into the network. This preprocessing step enhances the dataset's suitability for evaluating the effectiveness of combined learning methods within CSNNs, allowing for meaningful comparisons with previous studies. We then proceeded to assess the classification accuracy of the 1C-1S-FC architecture using the MNIST digit dataset, employing a combined learning approach with greedy layer-wise training. The results, depicted in Figure 6 and Figure 7 illustrate respectively the classification performance achieved by the 1C-1S-FC structure versus different kernel numbers and different architectures (1C-1S-FC, 2C-1S-FC, and 2C-1S-2C-FC) which demonstrates the efficacy of incorporating combined learning techniques into CSNN architectures. This underscores the potential of combined learning to enhance perceptual computing tasks, particularly when coupled with a layer-wise training strategy.

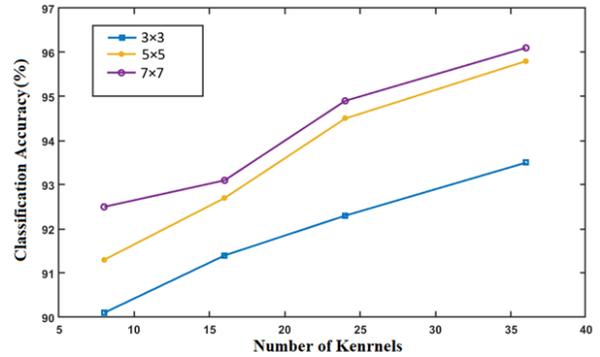

**Figure 6: Classification accuracy of the 1C-1S-FC for MNIST-digit dataset under combined -learning with greedy layer-wise training.**

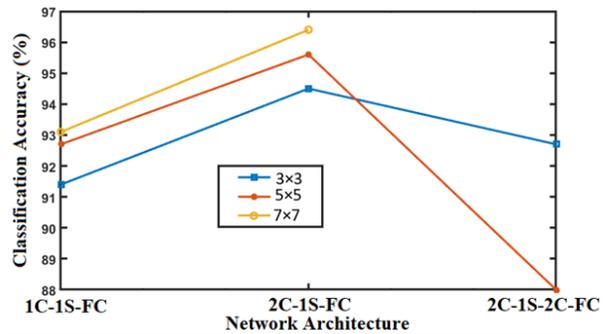

**Figure 7: Classification accuracy versus various CSNN architectures for different kernel sizes.**

The comparison of classification accuracy between recent studies and our proposed architecture, particularly focusing on the 1C-1S-FC configuration, demonstrates that the effectiveness of our approach in advancing perceptual computing systems has improved. This improvement is evident in Table 1. Through the integration of combined learning with greedy layer-wise training, we have attained notably superior accuracy rates when compared to existing methods [33]. This observation underscores the robustness and effectiveness of our architecture, particularly in handling complex perceptual computing tasks. In the context of digit classification tasks utilizing the MNIST dataset, our approach showcases remarkable performance improvements, as depicted in the comparison table presented in this study. The table offers a comprehensive overview of the classification accuracy achieved by our proposed architecture in comparison to previous works, emphasizing the significant strides made toward enhancing the accuracy and efficiency of perceptual computing systems.

Table 1: The comparison of the combined learning over SCNN with recent works for the MNIST digit dataset

| SNN topology | Architecture | Encoding Scheme | Learning Rule | Parameters | Accuracy |
|---|---|---|---|---|---|
| CSNN (our work) | Convolutional | Poisson-spike encoding | STDP using Combined learning | 23568 | 96.6% |
| CSNN [9] | Convolutional | Poisson-spike encoding | STDP | 25488 | 91.1% |
| CSNN [43] | Convolutional | Rank-order encoding | SVM+ STDP+ Sparse Coding | 590642 | 98.3% |
| SDNN [42] | Convolutional | Rank order encoding | STDP+ SVM | 76500 | 98.4% |
| Two-layer SNN [33] | Fully Connected | Poisson-spike encoding | STDP | 5017600 | 95% |

### 3.4 Test of Combined Learning with Caltech Images

To further analysis, the efficiency and robustness of the proposed combined learning method are evaluated through Caltech (Face/Motorbike) datasets. The categories of motorbikes and faces that cover two different objects have been selected. Two training sets are considered. Set one includes 200 images from the motorbike category, and set two includes 200 images from the face category. Both sets were randomly selected from a larger subset that contains 1276 images and the rest for testing sets. The elements of the dataset are converted to grayscale images and processed using the Laplacian of Gaussian (LoG) filter to signify the contrast of the pixels and image edges. Finally, the images of datasets are resized to dimension (40x40). The SCNN with three configurations and different kernel sizes is trained with training datasets. Figures 8 and 9 show the classification accuracy results for motorbike and face categories versus the various numbers of kernels and architectures, respectively. The classification accuracy of the proposed architecture for Caltech (Face/Motorbike) is compared with recent works shown in Table 2.

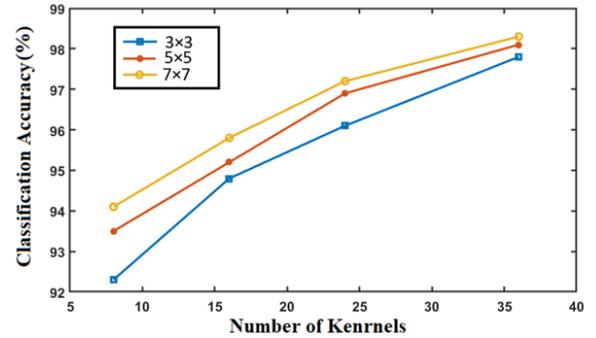

Figure 8: SCNN architectures for different kernel sizes for caltech dataset

The processed input pattern encoded using the Poisson spike train, the maximum frequency of the Poisson spike in simulations is 500 Hz, and each input pixel propagates the spike with the scaled frequency proportional to the normalized pixel intensity. The initial synaptic weights are selected randomly from a uniform distribution. The post-neurons integrate the input spike as the membrane threshold voltage, and if the membrane voltage exceeds a specific threshold, the post-neuron fires. The post-synaptic spikes' event updates synaptic efficacies of all the synapses connected to the firing post-neuron.

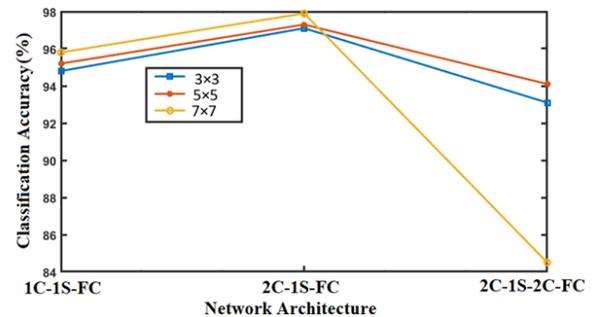

Figure 9: classification accuracy versus several network architectures for Caltech images

In the training of the SCNN, the synaptic weights that connect the convolutional layers to spatial pooling layer in 1C-1S-1FC and 2C-1S-FC configurations are not trainable. The connected synapses to final layer connections are trained using supervised STDP learning. Each final post-neuron is specialized to learn the specific pattern using threshold voltage adjustment of the neuron. The threshold voltage adjustment method controls the neurons' firing rate by an increment of threshold voltage and allowing it to decay exponentially. Each pattern is introduced for 25 epochs, including submission of pattern for 50 mS.

Table 2: The comparison of the combined learning over SCNN with recent works for the Caltech dataset

| SNN topology | Architecture | Encoding Scheme | Learning Rule | Parameters | Accuracy |
|---|---|---|---|---|---|
| SCNN (our work) | Convolutional | Poisson-spike encoding | STDP | 22572 | 98.3% |
| SCNN [9] | Convolutional | Poisson-spike encoding | STDP | 25488 | 97.6% |
| SCNN [44] | Convolutional | Rank-order encoding | Reinforcement STDP | 23120 | 98.9% |
| ScNN [45] | Convolutional | Rank order encoding | STDP+RBF | 20480 | 97.7% |
| SDNN [42] | Convolutional | Rank-order encoding | STDP+SVM | 25480 | 99.1% |

In neuromorphic hardware design, some machine learning and encoding techniques such as Support Vector Machine (SVM), Radial Basis Function, and Rank-order encoding must be implemented digitally and perform off-the-shelf training that burdens huge peripheral CMOS circuits and high power consumption. The combined learning technique enhances the recognition rate compared to some recent works, specifically with the proposed architecture

## 4 CONCLUSION

This study investigates the fusion of spiking neurons with convolutional neural networks through a synergistic learning strategy. This approach is evaluated using a combination of pair-based PSTDP and power-law-dependent STDP. The techniques have yielded impressive outcomes in training CSNN for the recognition and classification tasks on the MNIST and CALTECH datasets (Faces and Motorbikes), achieving this with a reduced count of trainable parameters relative to existing benchmarks. This reduction leads to lower energy consumption and minimizes the physical footprint by harnessing the stochastic behavior of memristive devices and the temporal dynamics of spiking neurons, the combined learning approach holds promise for implementing a crossbar structure using at least two parallel memristors as an entangled device.